\documentclass{article}
\usepackage{spconf,amsmath,graphicx}

\usepackage{enumitem}
\usepackage{subfigure}
\usepackage{booktabs}
\setlist{nosep, leftmargin=14pt}
\usepackage{balance}

\usepackage{mwe} 
\usepackage{threeparttable}
\usepackage{multicol}
\usepackage[skip=2pt]{caption}
\usepackage{placeins}
\usepackage[T1]{fontenc}
\usepackage[utf8]{inputenc}

\begin{document}
\twocolumn[{%
\begin{center}
\vspace{0.8em}
{\large\bfseries \MakeUppercase{Multi-Stage Fine-Tuning of Pathology Foundation Models with\\ Head-Diverse Ensembling for White Blood Cell Classification}\par}
\vspace{1.2em}
{\normalsize
Antony Gitau$^{1,2}$ \quad Martin Paulson$^{2}$ \quad Bj{\o}rn-Jostein Singstad$^{1,2}$ \\
Karl Thomas Hjelmervik$^{1}$ \quad Ola Marius Lysaker$^{1}$ \quad Veralia Gabriela Sanchez$^{1}$
}\\
\vspace{0.8em}
$^{1}$ University of South-Eastern Norway \\
$^{2}$ Vestfold Hospital Trust
\end{center}
\vspace{0.5em}
}]
\begin{abstract}

The classification of white blood cells (WBCs) from peripheral blood smears is critical for the diagnosis of leukemia. However, automated approaches still struggle due to challenges including class imbalance, domain shift, and morphological continuum confusion, where adjacent maturation stages exhibit subtle, overlapping features. We present a multi-stage fine-tuning methodology for 13-class WBC classification in the WBCBench 2026 Challenge (ISBI 2026). Our best-performing model is a fine-tuned DINOBloom-base, on which we train multiple classifier head families (linear, cosine, and multilayer perceptron (MLP)). The cosine head performed best on the mature granulocyte boundary (Band neutrophil (BNE) F1 = 0.470), the linear head on more immature granulocyte classes (Metamyelocyte (MMY) F1 = 0.585), and the MLP head on the most immature granulocyte (Promyelocyte (PMY) F1 = 0.733), revealing class-specific specialization. Based on this specialization, we construct a head-diverse ensemble, where the MLP head acts as the primary predictor, and its predictions within the four predefined confusion pairs are replaced only when two other head families agree. We further show that cases consistently misclassified by all models are substantially enriched for probable labeling errors or inherent morphological ambiguity.

\end{abstract}
\section{Introduction}
\label{sec:intro}
\begin{figure}[t]
  \centering
  \includegraphics[width=\linewidth]{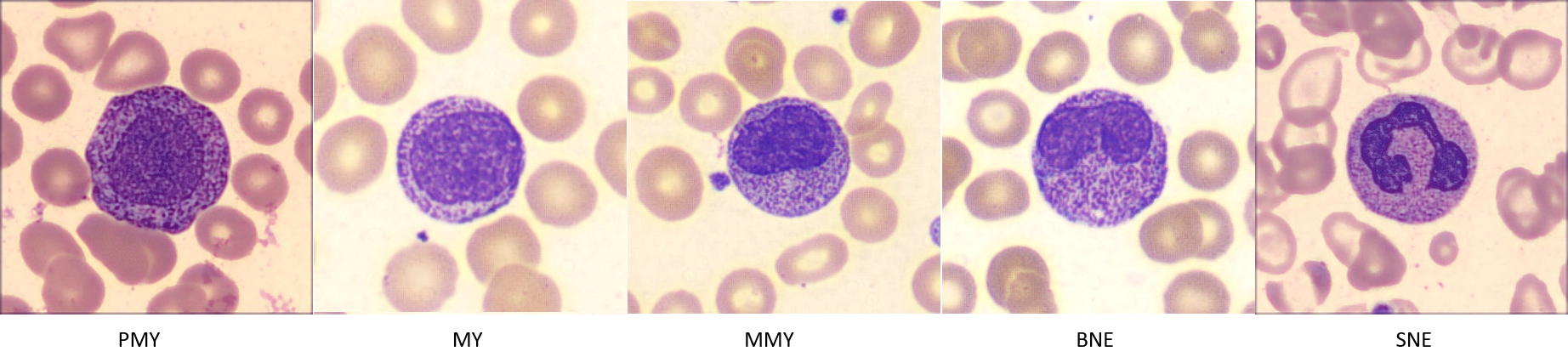}
  \caption{An illustration of the morphological continuum from promyelocyte (PMY), to myelocyte (MY), metamyelocyte (MMY), band-form neutrophil (BNE) and segmented neutrophil (SNE)}
  \label{fig:morphological_continuum}
\end{figure}

Microscopic examination of peripheral blood smears is clinically essential for diagnosing hematologic malignancies, such as leukemia, as well as non-malignant conditions, including infections and immune disorders. However, automated WBC classification remains challenging due to several factors:  severe class imbalance, where rare morphologies such as prolymphocytes (PLY) are underrepresented in normal peripheral blood \cite{Palmer2015ICSH}; morphological ambiguity arising from the continuum of cell maturation, where adjacent developmental stages exhibit subtle visual differences as shown in Figure ~\ref{fig:morphological_continuum}; and variability in imaging conditions across scanners and acquisition noise.

Recent pathology foundation models (FMs), pretrained on millions of histopathology and cytomorphology images, offer rich feature representations for downstream tasks. DinoBloom \cite{Koc_DinoBloom_MICCAI2024}, trained on over 380,000 single cell images from blood and bone marrow smears, provides domain-specific representations for hematology. However, standard linear probing, where a linear classifier is trained on features of frozen FM, may fail to fully utilize these representations for fine-grained classification under severe class imbalance.

Our contributions are summarized as follows;
\begin{itemize}
    \item We present a reproducible fine-tuning strategy for pathology FMs, with accompanying code that enables systematic backbone and head-family ablations to study the effect of different model configurations.
    
    \item We show that the geometry of the classifier head drives class-specific specialization within a common FM embedding space, motivating a head-diverse ensemble for resolving confusions along morphological continua.

    \item We analyze the WBCBench dataset, showing that many difficult cases may reflect either labeling errors or inherent morphological ambiguity.
\end{itemize}

\section{Materials and Methods}
\label{sec:format}

\begin{figure*}[t]
\centering
\includegraphics[width=\textwidth]{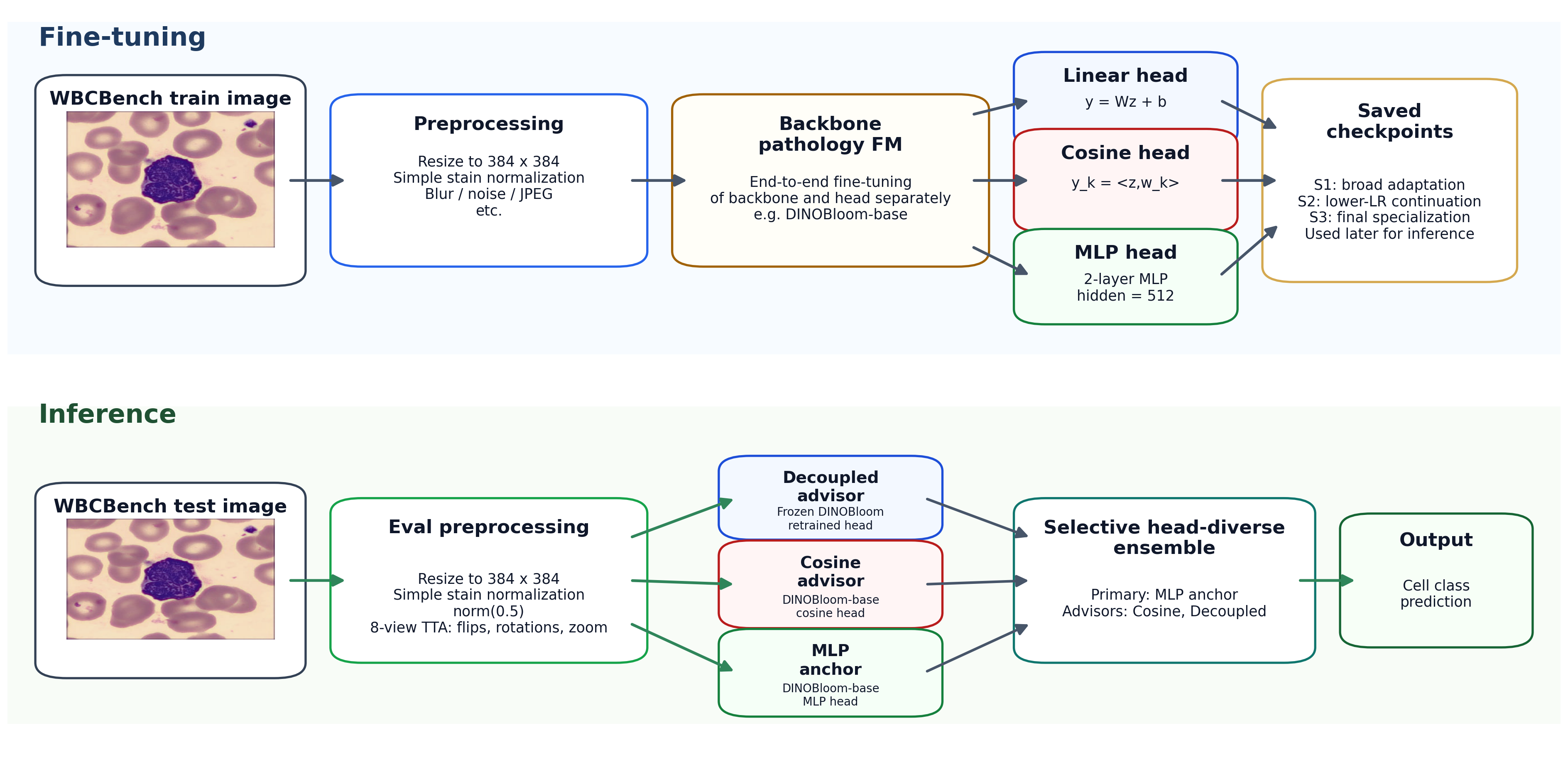}
\caption{End-to-end fine-tuning and inference simplified visual illustration. During training, separate models are obtained by fine-tuning a pathology foundation model (DINOBloom) with different classifier heads (linear, cosine, and MLP) across staged optimization. During inference, saved full DinoBloom-base and head checkpoints are combined using a head-diverse ensemble, where an MLP head acts as the primary predictor and is conditionally overridden by agreement between auxiliary heads.}
\label{fig:architecture}
\end{figure*}

\subsection{Dataset}
\begin{figure}[hp]
  \centering
  \includegraphics[width=\columnwidth]{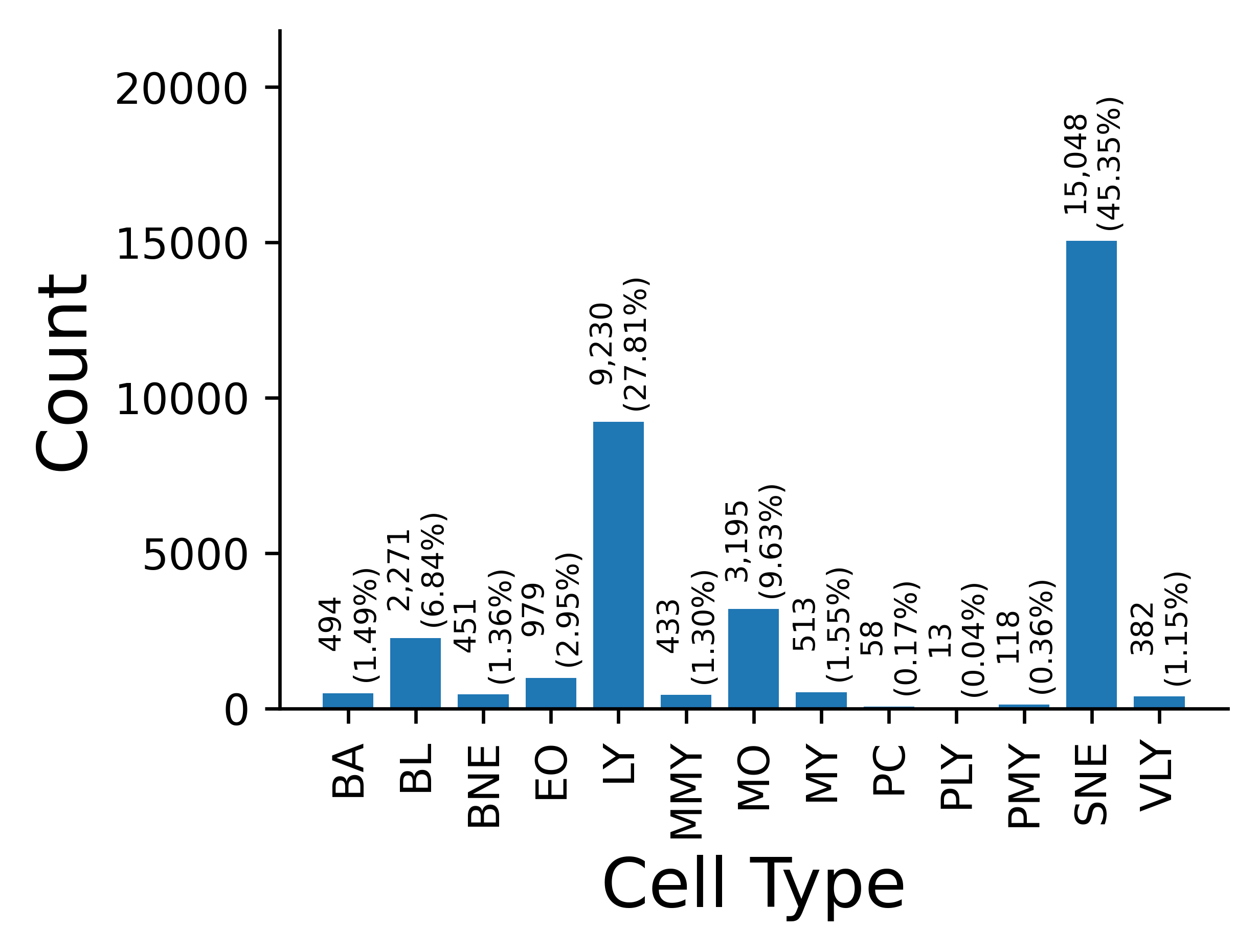}
  \begin{tablenotes}[center]
\footnotesize
\begin{multicols}{2}
\item[a] SNE: segmented neutrophil
\item[b] LY: lymphocyte
\item[c] MO: monocyte
\item[d] EO: eosinophil
\item[e] BA: basophil
\item[f] VLY: variant lymphocyte
\item[g] BNE: band-form neutrophil
\item[h] MMY: metamyelocyte
\item[i] MY: myelocyte
\item[j] PMY: promyelocyte
\item[k] BL: blast
\item[l] PC: plasma cell
\item[m] PLY: prolymphocyte
\end{multicols}
\end{tablenotes}
  \caption{Distributions of white blood cell types in the training set.}
  \label{fig:class-dist}
\end{figure}

We employed the official WBCBench 2026 challenge dataset~\cite{wbc-bench-2026}, comprising 55,012 images distributed across 13 classes, with splits of 60\% for training, 10\% for validation, and 30\% for testing. Figure~\ref{fig:class-dist} illustrates the class distribution in the training set.  
All images were first resized to $384 \times 384$ pixels and stain-normalized prior to applying augmentation. Training augmentations included geometric and color perturbations, along with quality degradations such as JPEG compression, resolution scaling, blur, and additive noise to improve robustness to test-time corruption. During inference, we applied 8-view test-time augmentation with logit averaging. Full pre-processing and augmentation details are provided in the accompanying code~\footnote{\texttt{github.com/Antony-gitau/msfit}}.

\subsection{Backbone Models and Classification Heads}

We evaluated multiple backbone families to balance hematology-specific pretraining and architectural diversity; the full set of explored backbones is documented in the accompanying codebase. For each backbone, let $f \in R^d$ denote the final feature vector produced by the backbone. 
\[
z = \mathrm{Dropout}(\mathrm{LayerNorm}(f)).
\]

We then apply a shared head. We evaluated three classification head families for $C=13$ classes:

\textbf{Linear head:}
\[
y = Wz + b
\]

\textbf{Cosine head:}
\[
y_k = \left\langle \frac{z}{\|z\|_2}, \frac{w_k}{\|w_k\|_2} \right\rangle
\]

\textbf{MLP head:}
\[
\begin{aligned}
h &= \mathrm{Dropout}(\mathrm{GELU}(W_1 z + b_1)), \\
y &= W_2 h + b_2
\end{aligned}
\]

\noindent The overall system design is visually presented in Figure \ref{fig:architecture}.

\subsection{Multi-Stage Fine-tuning}

We used a three-stage end-to-end fine-tuning schedule, where each stage initializes from the best checkpoint based on the validation F1 score of the previous stage. For both cosine and MLP head models, all backbone layers remained unfrozen throughout training.

Stage S1 was trained for 11 epochs using learning rates of $2.5 \times 10^{-5}$ (head) and $5.0 \times 10^{-6}$ (backbone) with 2 warmup epochs. Stage S2 continued from S1 for 5 epochs with reduced learning rates of $1.0 \times 10^{-5}$ and $2.0 \times 10^{-6}$, respectively. Stage S3 further refined the model for 5 epochs using learning rates of $5.0 \times 10^{-6}$ and $1.0 \times 10^{-6}$. The cosine-head models used the identical 11/5/5 epoch staging scheme that was previously applied with proportionally scaled learning rates.

For the fully fine-tuned MLP, linear, and cosine models, we used AdamW with automatic mixed precision and gradient accumulation. These models were trained using focal loss ($\gamma = 2.0$), label smoothing ($\epsilon = 0.1$). We used class-dependent focal weights $\alpha_c \propto 1/\sqrt{n_c}$, normalized across classes, where $n_c$ denotes the number of training samples in class $c$ and MixUp/CutMix ($p = 0.1$ each). In addition, we trained a frozen-backbone MLP model initialized from the MLP-S3 checkpoint,  following the decoupled training idea of \cite{kang2020decoupling}, using cross-entropy with effective-number class weights and a balanced sampler. We also monitored a tail-aware composite metric where TailMacroF1 is assessed over the set of classes ({ $\mathrm{BNE, PLY, VLY, MMY, PMY, MY, PC}$ }).
 \[
 \mathrm{TailComposite}=\tfrac12,\mathrm{MacroF1}+\tfrac12,\mathrm{TailMacroF1},
 \]

\subsection{Head-Diverse Ensembling}
During inference, we employ head-diverse ensembling to combine complementary decision boundaries from different classifier heads. The fully fine-tuned MLP model serves as the primary predictor, while the cosine head model and the frozen-backbone MLP head model act as advisors. This means that if both of them disagree with the primary predictor, the output will be replaced.  Although a linear head model achieved competitive performance, it was not included in the final ensemble due to lower complementarity. Rather than using global voting or weight optimization, we use a conservative agreement-gated override rule that allows modifications only for a small, pre-approved set of ordered confusion transitions:

Let $\mathcal{P}$ denote the validated set of ordered confusion pairs identified on the validation set:
\begin{equation}
\mathcal{P} =
\left\{
(\text{BNE}, \text{SNE}),
(\text{MO}, \text{VLY}),
(\text{MY}, \text{MMY}),
(\text{LY}, \text{BL})
\right\}.
\end{equation}

Each pair $(c_i, c_j)$ indicates that replacing the primary prediction $c_i$ with $c_j$ improved validation performance. Let $\hat{y}_{\text{primary}}$ denote the primary prediction, and
$\hat{y}_{a1}, \hat{y}_{a2}$ denote the predictions of Advisor~1 and Advisor~2, respectively.

The final prediction is defined as:
\begin{equation}
\hat{y} =
\begin{cases}
\hat{y}_{a1}, &
\text{if } 
\left( \hat{y}_{a1} = \hat{y}_{a2} \right)
\land
\left( (\hat{y}_{\text{primary}}, \hat{y}_{a1}) \in \mathcal{P} \right)
\\[6pt]
\hat{y}_{\text{primary}}, & \text{otherwise}.
\end{cases}
\end{equation}

This design preserves the stability of the primary model while allowing overrides only when both advisors agree on a disagreement pattern.

\subsection{Expert Review of Disagreement Cases}
To assess label reliability, we used predictions from a single end-to-end fine-tuned DINOBloom-base model with a linear classification head, selected as the strongest available single-model checkpoint at the time the expert review was conducted. For each image in the training and validation sets, we recorded the top-3 predicted classes, their softmax probabilities, and the confidence margin $m = p_1 - p_2$, where $p_1$ and $p_2$ denote the top-1 and top-2 predicted probabilities, respectively. Discordant cases were defined as samples for which the top-1 predicted class differed from the assigned reference label. Expert hematologists from Vestfold Hospital Trust reviewed all discordant cases. Each case was categorized as (i) \textit{Label Error}, when the model prediction was judged correct and the original label incorrect; (ii) \textit{Model Error}, when the original label was judged correct and the model prediction incorrect; or (iii) \textit{Ambiguous}, when both interpretations were considered defensible.  To assess label reliability beyond model--label disagreement, we additionally reviewed a class-stratified random sample of training images for which the model prediction agreed with the assigned label.


\section{Results}
\label{sec:pagestyle}
\begin{figure}[t]
  \centering
  \includegraphics[width=\columnwidth]{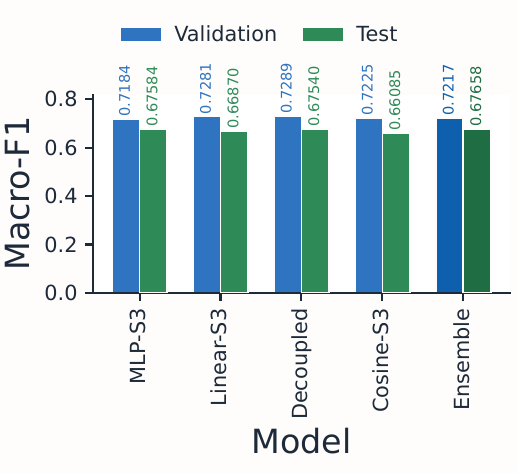}
  \caption{Comparison of various of our models' performances on the validation and test sets}
  \label{fig:eval_lb}
\end{figure}

\begin{figure}[t]
  \centering
  \includegraphics[width=\columnwidth]{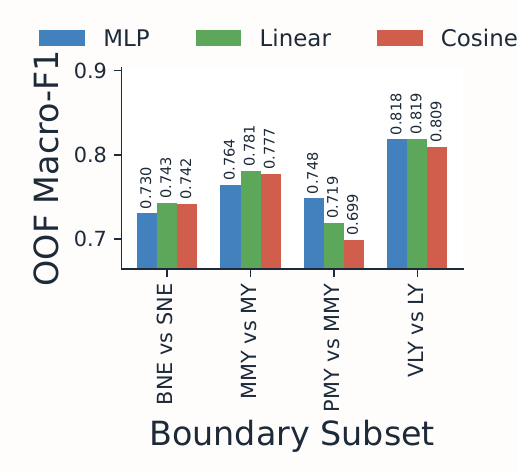}
  \caption{Comparison of classification head performance for the boundary subsets.}
  \label{fig:tail_family}
\end{figure}

Our best model was a head-diverse ensemble built on a fine-tuned DINOBloom-base backbone with three distinct heads as visualized in Figure \ref{fig:architecture}. On the organizer validation set, this ensemble achieved a Macro-F1 of 0.7217, and on the hidden test set, it reached 0.67658, reflecting an evaluation-to-test drop of roughly 0.045 in Macro-F1 score. As shown in Figure~\ref{fig:eval_lb}, this transfer gap was not unique to the ensemble. Higher macro F1 on the validation set did not reliably translate to stronger hidden-test performance. For example, linear and decoupled variants achieved higher validation Macro-F1 than the ensemble, yet both scored lower on the test set. 
To mitigate reliance on a particular data split, we perform a 5-fold out-of-fold analysis, training each head family on four folds and evaluating on the held-out fold before aggregating predictions across all folds. As shown in Figure~\ref{fig:tail_family}, boundary-level performance is head-dependent: linear and cosine are stronger than MLP on the difficult BNE-SNE boundary, linear is strongest on MMY-MY, and MLP is strongest on PMY-MMY. These trends are consistent with the provided validation set where cosine scored high on BNE, linear on MMY, and MLP on PMY.

\begin{figure}[!t]
  \centering
  \includegraphics[width=\columnwidth]{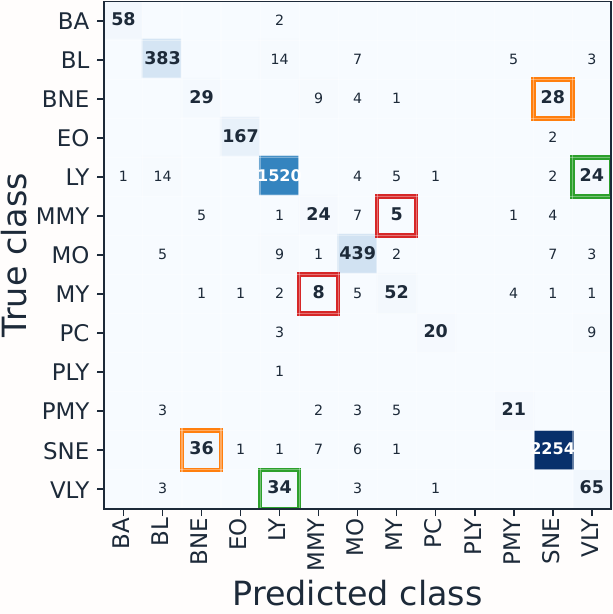}
  \caption{Evaluation confusion matrix of the final hybrid. Highlighted regions indicate recurring confusions relevant to the conservative override rule.}
  \label{fig:confusion_matrix}
\end{figure}

Relative to the primary MLP predictions, the final ensemble modified only 20 out of 16,477 test outputs (0.12\%), yet increased test performance by 0.00074. Figure \ref{fig:confusion_matrix} presents its confusion matrix, highlighting the recurrent error regions that motivated the restricted override strategy.

\subsection{Expert Label Review Results}
Expert review showed that model-label disagreement was enriched for annotation errors, as summarized in Table~\ref{tab:expert_label_review_summary}. Across the training and validation sets, 54.5\% of discordant cases were judged to be \textit{Label Error}, compared with 34.9\% \textit{Model Error} and 10.6\% \textit{Ambiguous}. Additionally, the reviewed 1,954 model-label agreement cases identified more mislabeled samples (8.0\%) and ambiguous cases (6.4\%), indicating that label noise was not confined to discordant predictions. Agreement-set review also suggested elevated per-class label noise in PC (25.5\%), LY (21.7\%), VLY (16.7\%), PMY (14.6\%), and MMY (12.8\%).

\begin{table}[h]
  \centering
  \footnotesize
  \setlength{\tabcolsep}{4pt}
  \begin{tabular}{lcccc}
  \toprule
  Disagreement & $n$ & Label Error & Model Error & Ambig. \\
  \midrule
  Train & 1,457 & 803 (55.1\%) & 484 (33.2\%) & 170 (11.7\%) \\
  Validation & 332 & 172 (51.8\%) & 140 (42.2\%) & 20 (6.0\%) \\
  Combined & 1,789 & 975 (54.5\%) & 624 (34.9\%) & 190 (10.6\%) \\
  \midrule
  Agreement & $n$ & Label Error & Ambig. & Confirmed Correct \\
  \midrule
  Train & 1,954 & 156 (8.0\%) & 125 (6.4\%) & 1,673 (85.6\%) \\
  \bottomrule
  \end{tabular}
  \caption{Summary of expert label review outcomes.}
  \label{tab:expert_label_review_summary}
  \end{table}

\begin{figure}[h]
  \centering
  \includegraphics[width=\columnwidth]{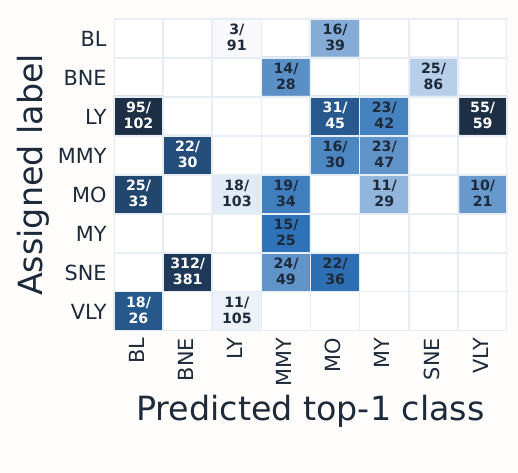}
  \caption{Directional analysis of reviewed disagreement cases across training and validation sets. Rows denote assigned
  labels and columns denote model top-1 predictions. Each cell shows label errors / reviewed disagreements for that pair,
  and color indicates the label-error rate..}
  \label{fig:heatmap}
\end{figure}

Directional analysis of discordant cases showed that label noise was highly asymmetric, as illustrated in Figure~\ref{fig:heatmap}. The strongest label-noise pairs were LY$\rightarrow$VLY, LY$\rightarrow$BL, and SNE$\rightarrow$BNE, whereas the reverse directions were often model errors, including BL$\rightarrow$LY, VLY$\rightarrow$LY, BNE$\rightarrow$SNE, and MO$\rightarrow$LY.

\section{Discussion and Conclusion}
\label{sec:typestyle}
We present a multi-stage fine-tuning approach for pathology FMs for the WBC classification task. Three main findings emerged from this study. First, classifier-head geometry should be matched to the morphology subgroup rather than selected globally, as no single classification head dominated across all difficult classes.

Second, under the observed validation-to-test shift, a head-diverse ensemble transferred better to the hidden test set than, for example, a broader 9-model weighted ensemble built from all three stages of the three DINOBloom head families (MLP, linear, and cosine). This broader ensemble achieved the highest validation Macro-F1 (~0.7400) but a lower leaderboard score (0.67285) than the head-diverse ensemble, which scored 0.7217 on validation macro F1 and a 0.67658 on the leaderboard.

Third, expert review showed that label noise was non-trivial and highly structured rather than uniformly distributed. More than half of the reviewed model--label disagreements were judged as label errors, and additional samples judged as mislabels were also found among agreement cases. Moreover, noise seemed concentrated in morphologically adjacent pairs, particularly SNE$\rightarrow$BNE and LY$\rightarrow$BL/VLY, whereas the reverse directions were often model errors. This indicates that model--label disagreement (and agreements) analysis can serve as an auditing signal for label noise.

 
 From a clinical workflow perspective, robust automated pre-classification may reduce manual microscopy burden, but high-stakes use still requires expert review and site-specific validation, especially for rare classes. A limitation of this study is the observed label noise, which is a known problem across medical imaging literature, and the extreme rarity of some classes, such as PLY, which constrains reliable learning and evaluation for such classes, hence limiting the utility of deep learning solutions in real-world clinical setups.  
 
 In addition, although the multi-stage fine-tuning protocol performed well in our experiments and transferred better to the hidden test set, we did not include a matched single-stage baseline with equivalent training budget. As a result, the staged design should be interpreted as a strong practical strategy rather than as a uniquely validated requirement.

\section{Compliance with Ethical Standards}
This research study was conducted retrospectively using human subject data made available in open access by Xin et al 2026~\cite{wbc-bench-2026}. Ethical approval was not required as confirmed by the license attached with the open access data

\bibliographystyle{IEEEbib}
\bibliography{strings,refs}
\end{document}